\documentclass[runningheads]{llncs}
\usepackage{graphicx,floatrow,booktabs}

\usepackage{hyperref}
\hypersetup{
    colorlinks=true,
    linkcolor=blue,
    filecolor=magenta,      
    urlcolor=cyan,
}

\usepackage{graphicx}
\usepackage[document]{ragged2e}
\begin{document}
\title{Exploiting Deep Learning for Persian Sentiment Analysis}
\author{Kia Dashtipour\inst{1} \and
Mandar Gogate\inst{1} \and
Ahsan Adeel\inst{1}\and
Cosimo Ieracitano\inst{2}
Hadi Larijani\inst{3}\and
Amir Hussain\inst{1}}
\authorrunning{K. Dashtipour et al.}

\institute{Department of Computing Science and Mathematics, Faculty of Natural Sciences, University of Stirling, FK9 4LA Stirling, United Kingdom \\
%\email{\{cosimo.ieracitano,morabito\}@unirc.it}\\
 \and
DICEAM Department, University Mediterranea of Reggio Calabria, 89124 Reggio Calabria, Italy\\
\and
Department of Communication, Network and Electronic Engineering, Glasgow Caledonian University, G4 0BA Glasgow, United Kingdom\\
\email{\inst{*}\ email:kd28@cs.stir.ac.uk}}

\maketitle              % typeset the header of the contribution
\begin{abstract}
The rise of social media is enabling people to freely express their opinions about products and services. The aim of sentiment analysis is to automatically determine subject's sentiment (e.g., positive, negative, or neutral) towards a particular aspect such as topic, product, movie, news etc. Deep learning has recently emerged as a powerful machine learning technique to tackle a growing demand of accurate sentiment analysis. However, limited work has been conducted to apply deep learning algorithms to languages other than English, such as Persian. In this work, two deep learning models (deep autoencoders and deep convolutional neural networks (CNNs)) are developed and applied to a novel Persian movie reviews dataset. The proposed deep learning models are analyzed and compared with the the state-of-the-art shallow multilayer perceptron (MLP) based machine learning model. Simulation results demonstrate the enhanced performance of deep learning over state-of-the-art MLP.

\end{abstract}
\section{Introduction}

In recent years, social media, forums, blogs and other forms of online communication tools have radically affected everyday life, especially how people express their opinions and comments. The extraction of useful information (such as people's opinion about companies brand) from the huge amount of unstructured data is vital for most companies and organizations\cite{dashtipour2017persianA}. The product reviews are important for business owners as they can take business decision accordingly to automatically classify user’s opinions towards products and services. The application of sentiment analysis is not limited to product or movie reviews but can be applied to different fields such as news, politics, sport etc. For example, in online political debates, the sentiment analysis can be used to identify people's opinions on a certain election candidate or political parties \cite{tan2017mining}\cite{ren2009hierarchical}\cite{ren2010activity}. In this context, sentiment analysis has been widely used in different languages by using traditional and advanced machine learning techniques. However, limited research has been conducted to develop models for the Persian language. 

The sentiment analysis is a method to automatically process large amounts of data and classify text into positive or negative sentiments) \cite{cambria2018senticnet} \cite{dashtipour2016multilingual}. Sentiment analysis can be performed at two levels: at the document level or at the sentence level. At document level it is used to classify the sentiment expressed in the document (positive or negative), whereas, at sentence level is used to identify the sentiments expressed only in the sentence under analysis \cite{dashtipour2016persent} \cite{dashtipour2017persian}. 

In the literature, deep learning based automated feature extraction has been shown to outperform state-of-the-art manual feature engineering based classifiers such as Support Vector Machine (SVM), Naive Bayes (NB) or Multilayer Perceptron (MLP)  etc. One of the important techniques in deep learning is the autoencoder that generally involves reducing the number of feature dimensions under consideration. The aim of dimensionality reduction is to obtain a set of principal variables to improve the performance of the approach. Similarly, CNNs have been proven to be very effective in sentiment analysis. However, little work has been carried out to exploit deep learning based feature representation for Persian sentiment analysis \cite{lecun2015deep} \cite{gasparini2018information}. In this paper, we present two deep learning models (deep autoencoders and CNNs) for Persian sentiment analysis. The obtained deep learning results are compared with MLP.

The rest of the paper is organized as follows: Section 2 presents related work. Section 3 presents methodology and experimental results. Finally, section 4 concludes this paper.

\section{Related Works}
In the literature, extensive research has been carried out to model novel sentiment analysis models using both shallow and deep learning algorithms. For example, the authors in \cite{chen2012marginalized} proposed a novel deep learning approach for polarity detection in product reviews. The authors addressed two major limitations of stacked denoising of autoencoders, high computational cost and the lack of scalability of high dimensional features. Their experimental results showed the effectiveness of proposed autoencoders in achieving accuracy upto 87\%. Zhai et al., \cite{zhai2016semisupervised} proposed a five layers autoencoder for learning the specific representation of textual data. The autoencoders are generalised using loss function and derived discriminative loss function from label information. The experimental results showed that the model outperformed bag of words, denoising autoencoders and other traditional methods, achieving accuracy rate up to 85\% . Sun et al., \cite{sun2016sentiment} proposed a novel method to extract contextual information from text using a convolutional autoencoder architecture. The experimental results showed that the proposed model outperformed traditional SVM and Nave Bayes models, reporting accuracy of 83.1 \%, 63.9\% and 67.8\% respectively. 

Su et al., \cite{su2018neural} proposed an approach for a neural generative autoencoder for learning bilingual word embedding. The experimental results showed the effectiveness of their approach on English-Chinese, English-German, English-French and English-Spanish (75.36\% accuracy).  Kim et al., \cite{kim2014convolutional} proposed a method to capture the non-linear structure of data using CNN classifier. The experimental results showed the effectiveness of the method on the multi-domain dataset (movie reviews and product reviews). However, the disadvantage is only SVM and Naive Bayes classifiers are used to evaluate the performance of the method and deep learning classifiers are not exploited. Zhang et al., \cite{zhang2015japanese} proposed an approach using deep learning classifiers to detect polarity in Japanese movie reviews. The approach used denoising autoencoder and adapted to other domains such as product reviews. The advantage of the approach is not depended on any language and could be used for various languages by applying different datasets. AP et al., \cite{ap2014autoencoder} proposed a CNN based model for cross-language learning of vectorial word representations that is coherent between two languages. The method is evaluated using English and German  movie reviews dataset. The experimental results showed CNN (83.45\% accuracy) outperformed as compared to SVM (65.25\% accuracy). 

Zhou et al., \cite{zhou2015learning} proposed an autoencoder architecture constituting an LSTM-encoder and decoder in order to capture features in the text and reduce dimensionality of data. The LSTM encoder used the interactive scheme to go through the sequence of sentences and LSTM decoder reconstructed the vector of sentences. The model is evaluated using different datasets such as book reviews, DVD reviews, and music reviews, acquiring accuracy up to 81.05\%, 81.06\%, and 79.40\% respectively. Mesnil et al., \cite{mesnil2014ensemble} proposed an approach using ensemble classification to detect polarity in the movie reviews. The authors combined several machine learning algorithms such as SVM, Naive Bayes and RNN to achieve better results, where autoencoders were used to reduce the dimensionality of features. The experimental results showed the combination of unigram, bigram and trigram features (91.87\% accuracy) outperformed unigram (91.56\% accuracy) and bigram (88.61\% accuracy).

Scheible et al., \cite{scheible2013cutting} trained an approach using semi-supervised recursive autoencoder to detect polarity in movie reviews dataset, consisted of 5000 positive and 5000 negative sentiments. The experimental results demonstrated that the proposed approach successfully detected polarity in movie reviews dataset (83.13\% accuracy) and outperformed standard SVM (68.36\% accuracy) model. Dai et al., \cite{dai2015semi} developed an autoencoder to detect polarity in the text using deep learning classifier. The LSTM was trained on IMDB movie reviews dataset. The experimental results showed the outperformance of their proposed approach over SVM. In table 1 some of the autoencoder approaches are depicted.

\section{Methodology  and Experimental Results}

The novel dataset used in this work was collected manually and includes Persian movie reviews from 2014 to 2016. A subset of dataset was used to train the neural network (60\% training dataset) and rest of the data (40\%) was used to test and validate the performance of the trained neural network (testing set (30\%), validation set (10\%)). There are two types of labels in the dataset: positive or negative. The reviews were manually annotated by three native Persian speakers aged between 30 and 50 years old. 

After data collection, the corpus was pre-processed using tokenisation, normalisation and stemming techniques. The process of converting sentences into single word or token is called tokenisation. For example, "The movie is great" is changed to "The", "movie", "is", "great" \cite{sumathy2013text}. There are some words which contain numbers. For example, "great" is written as "gr8" or "gooood" as written as "good" . The normalisation is used to convert these words into normal forms \cite{reynolds1997comparison}. The process of converting words into their root is called stemming. For example, going was changed to go \cite{korenius2004stemming}. Words were converted into vectors. The fasttext was used to convert each word into 300-dimensions vectors. Fasttext is a library for text classification and representation \cite{joulin2016fasttext}\cite{morabito2016deep}\cite{gasparini2018information}. 

For classification, MLP, autoencoders and CNNs have been used. Fig. 1. depicts the modelled MLP architectures. MLP classifer was trained for 100 iterations \cite{gardner1998artificial}. Fig. 2. depicts the modelled autoencoder architecture. Autoencoder is a feed-forward deep neural network with unsupervised learning and it is used for dimensionality reduction. The autoencoder consists of input, output and hidden layers. Autoencoder is used to compress the input into a latent-space and then the output is reconstructed \cite{semeniuta2017hybrid} \cite{gogate2017deep} \cite{gogate2017novel}. The exploited autoencoder model is depcited in Fig. 1. The autoencoder consists of one input layer three hidden layers (1500, 512, 1500) and an output layer. Convolutional Neural Networks contains three layers (input, hidden and output layer). The hidden layer consists of convolutional layers, pooling layers, fully connected layers and normalisation layer. The  \( h_{j} \)  is denotes the hidden neurons of j, with bias of \( ~b_{j} \) , is a weight sum over continuous visible nodes v which is given by:\par

\begin{equation}
h_{j}=~b_{j}+~ \sum ~v_{i}w_{ij} 
\end{equation}

The modelled CNN architecture is depicted in Fig. 3 \cite{gogate2017novel}\cite{gogate2017deep}. For CNN modelling, each utterance was represented as a concatenation vector of constituent words. The network has total 11 layers: 4 convolution layers, 4 max pooling and 3 fully connected layers. Convolution layers have filters of size 2 and with 15 feature maps. Each convolution layer is followed by a max polling layer with window size 2. The last max pooling layer is followed by fully connected layers of size 5000, 500 and 4. For final layer, softmax activation is used.

\begin{figure}[h]
\centering
\includegraphics[width=5cm, height=5cm]{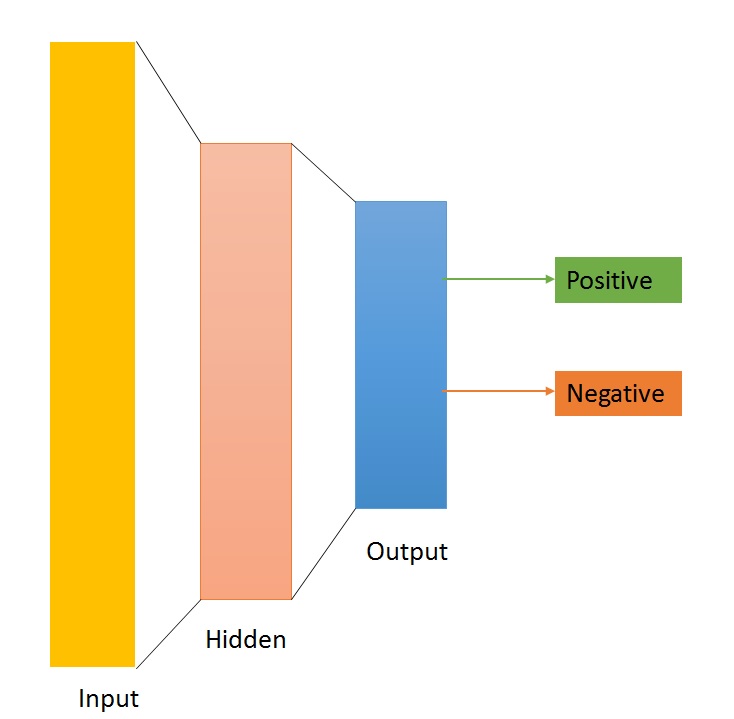}
\caption {Multilayer Perceptron} 
\end{figure}

%%%%%%%%%%%%%%%%% Figure/Image No: 2 starts %%%%%%%%%%%%%%%%%%%%

\begin{figure}[h]
\centering
\includegraphics[width=6cm, height=5cm]{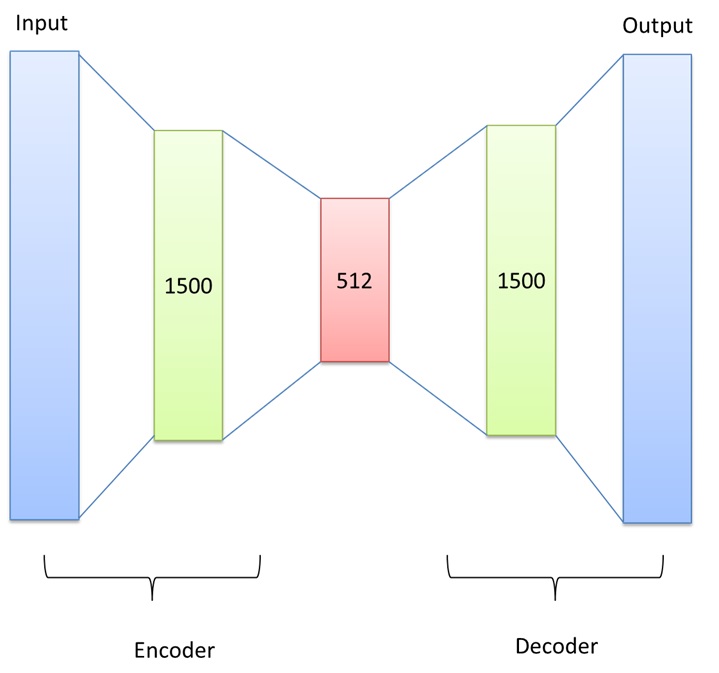}
\caption {Autoencoder} 
\end{figure}
%%%%%%%%%%%%%% Figure/Image No: 2 Ends here%%%%%%%%%%%%%%%%%%%%

%%%%%%%%%%%%%%%%%%%% Figure/Image No: 1 starts %%%%%%%%%%%%%%%%%%%%

\begin{figure}[h]
\centering
\includegraphics[width=12cm, height=7cm]{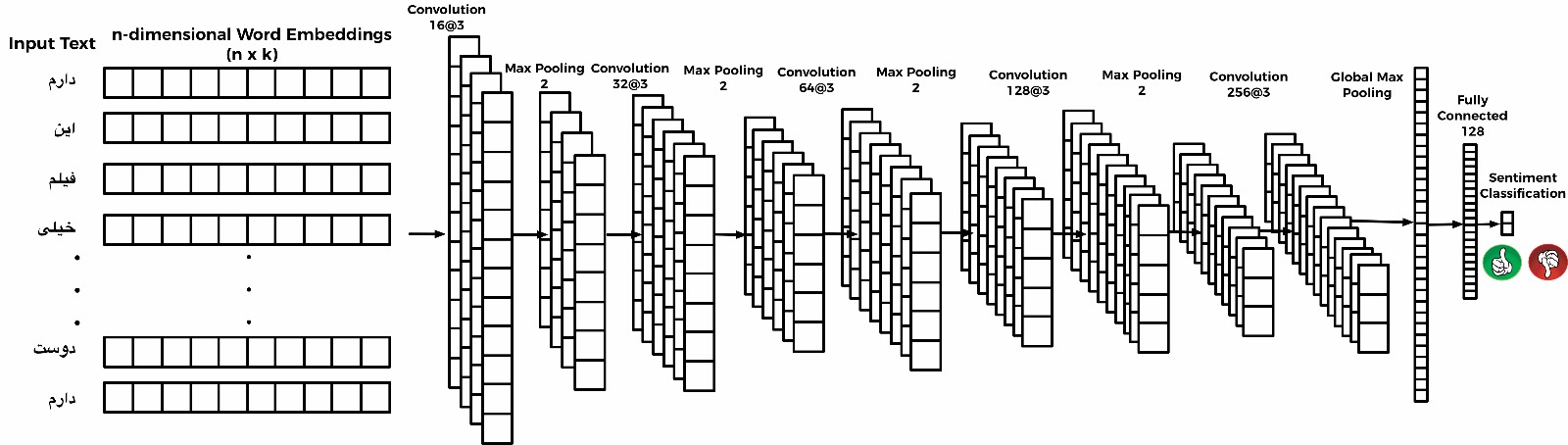}
\caption {Deep Convolutional Neural Network} 
\end{figure}

\break

To evaluate the performance of the proposed approach, precision (1), recall (2), f-Measure (3), and prediction accuracy (4) have been used as a performance matrices. The experimental results  are shown in Table 1, where it can be seen that autoencoders outperformed MLP and CNN outperformed autoencoders with the highest achieved accuracy of 82.6\%.

\begin{equation}
Precision =\frac{TP}{TP+FP}
\end{equation}
\begin{equation}
Recall =\frac{TP}{TP+FN}
\end{equation}
\begin{equation}
F\_measure =2*\frac{Precision*Recall}{Precision+Recall}
\end{equation}
\begin{equation}
Accuracy =\frac{TP+TN}{TP+TN+FP+FN}
\end{equation}

%%%%%%%%%%%%%%%%%Table 1%%%%%%%%%%%%%%%%%%%%%%%%%%%%%%%%%%

where TP is denotes true positive, TN is true negative, FP is false positive, and FN is false negative. 
% \usepackage{multirow}
\begin{table}[h]
\centering
\caption{Results: MLP vs. Autoencoder vs. Convolutional Neural Network}
\label{my-label}
\begin{tabular}{|l|l|l|l|l|}
\hline
\multicolumn{5}{|c|}{\textbf{MLP}}                                                        \\ \hline
          & \textbf{Precision} & \textbf{Recall} & \textbf{F-measure} & \textbf{Accuracy} (\%) \\ \hline
Negative  & 0.78               & 0.76            & 0.77               &  \\ \cline{1-4}
Positive  & 0.79               & 0.81            & 0.8                &                   \\ \hline
AVG       & 0.78               & 0.78            & 0.78               & 78.49             \\ \hline
\multicolumn{5}{|c|}{\textbf{MLP-Autoencoder}}                                            \\ \hline
          & \textbf{Precision} & \textbf{Recall} & \textbf{F-measure} & \textbf{Accuracy} (\%) \\ \hline
Negative  & 0.78               & 0.81            & 0.79               &  \\ \cline{1-4}
Positive  & 0.82               & 0.8             & 0.81               &                   \\ \hline
AVG       & 0.8                & 0.8             & 0.8                & 80.08             \\ \hline
\multicolumn{5}{|c|}{\textbf{1D-CNN}}                                                     \\ \hline
\textbf{} & \textbf{Precision} & \textbf{Recall} & \textbf{F-measure} & \textbf{Accuracy} (\%) \\ \hline
Negative  & 0.90               & 0.78            & 0.83               &  \\ \cline{1-4}
Positive  & 0.77               & 0.89            & 0.82               &                   \\ \hline
AVG       & 0.84               & 0.83            & 0.83               & 82.86             \\ \hline
\end{tabular}
\end{table}

\section {Conclusion}
Sentiment analysis has been used extensively for a wide of range of real-world applications, ranging from product reviews, surveys feedback, to business intelligence, and operational improvements. However, the majority of research efforts are devoted to English-language only, where information of great importance is also available in other languages. In this work, we focus on developing sentiment analysis models for Persian language, specifically for Persian movie reviews. Two deep learning models (deep autoencoders and deep CNNs) are developed and compared with the the state-of-the-art shallow MLP based machine learning model. Simulations results revealed the outperformance of our proposed CNN model over autoencoders and MLP. In future, we intend to exploit more advanced deep learning models such as Long Short-Term Memory (LSTM) and LSTM-CNNs to further evaluate the performance of our developed novel Persian dataset.

\section{Acknowledgment}
Amir Hussain and Ahsan Adeel were supported by the UK Engineering and Physical Sciences Research Council (EPSRC) grant No.EP/M026981/1.

\vspace{25em}
\break

\bibliographystyle{splncs04}
\bibliography{references}

\end{document}